\documentclass{article}

\usepackage[main,final,nonatbib]{neurips_2026}

\usepackage[utf8]{inputenc} 
\usepackage[T1]{fontenc}    
\usepackage{hyperref}       
\usepackage{url}            
\usepackage{booktabs}       
\usepackage{amsfonts}       
\usepackage{nicefrac}       
\usepackage{microtype}      
\usepackage{xcolor}         
\usepackage{graphicx}
\usepackage{wrapfig}
\usepackage{amsmath}
\usepackage{makecell}
\usepackage{multirow}   

\title{Manifold Prior Guided Deep Unfolding for Hyperspectral Image Reconstruction}

\author{%
  \textbf{Shuqi Liu}{\mdseries\textsuperscript{1}} \quad
  \textbf{Yingkai Zhang}{\mdseries\textsuperscript{1}} \quad
  \textbf{Lin Gu}{\mdseries\textsuperscript{2}}\\
  \textbf{Tao Zhang}{\mdseries\textsuperscript{3}} \quad
  \textbf{Ying Fu}{\mdseries\textsuperscript{1}}\thanks{Corresponding author.}\\[2pt]
  {\mdseries\textsuperscript{1}Beijing Institute of Technology}\\
  {\mdseries\textsuperscript{2}Tohoku University}\\
  {\mdseries\textsuperscript{3}Hangzhou Dianzi University}\\
  {\mdseries\texttt{\{shuqiliu,zhangyingkai,fuying\}@bit.edu.cn}}\\
  {\mdseries\texttt{lin@tohoku.ac.jp} \quad \texttt{tzhang@hdu.edu.cn}}
}

\begin{document}
\maketitle

\begin{abstract}
Hyperspectral Image (HSI) reconstruction has made substantial progress with the deep unfolding framework by decomposing the problem into a projection module and a denoiser module.
Nevertheless, existing methods still exhibit limitations of insufficient matching with HSI data. The issues lie in two aspects: 1) the data-consistency projection module applies a fixed gradient descent step while ignoring the spectral content adaptivity of HSI; 2) the denoiser module either suffers from high computational complexity or is limited to the local receptive field, failing to efficiently capture the global spatial-spectral dependencies.
In this paper, we propose a deep unfolding framework centered on a two-stage manifold learning strategy to obtain a degradation-free structural prior, which is then strategically injected into both the projection and denoiser modules.
In the data-consistency projection, a prior-guided spectral weight mechanism is introduced to adaptively modulate the weight of spectral band. 
In the denoiser module, a manifold-guided Mamba is proposed to achieve the structure-aware extraction of spatial-spectral features through prior modulation. 
Quantitative and visual comparisons on synthetic and real-world datasets show that the proposed method significantly outperforms existing state-of-the-art approaches, while maintaining the structural interpretability of the unfolding framework. The code is available at \url{https://github.com/SqLiu67/MPG-DU}.

\end{abstract}

\section{Introduction}

\begin{wrapfigure}[10]{r}{0.40\textwidth}
    \centering
    \vspace{-6pt}
    \includegraphics[width=\linewidth]{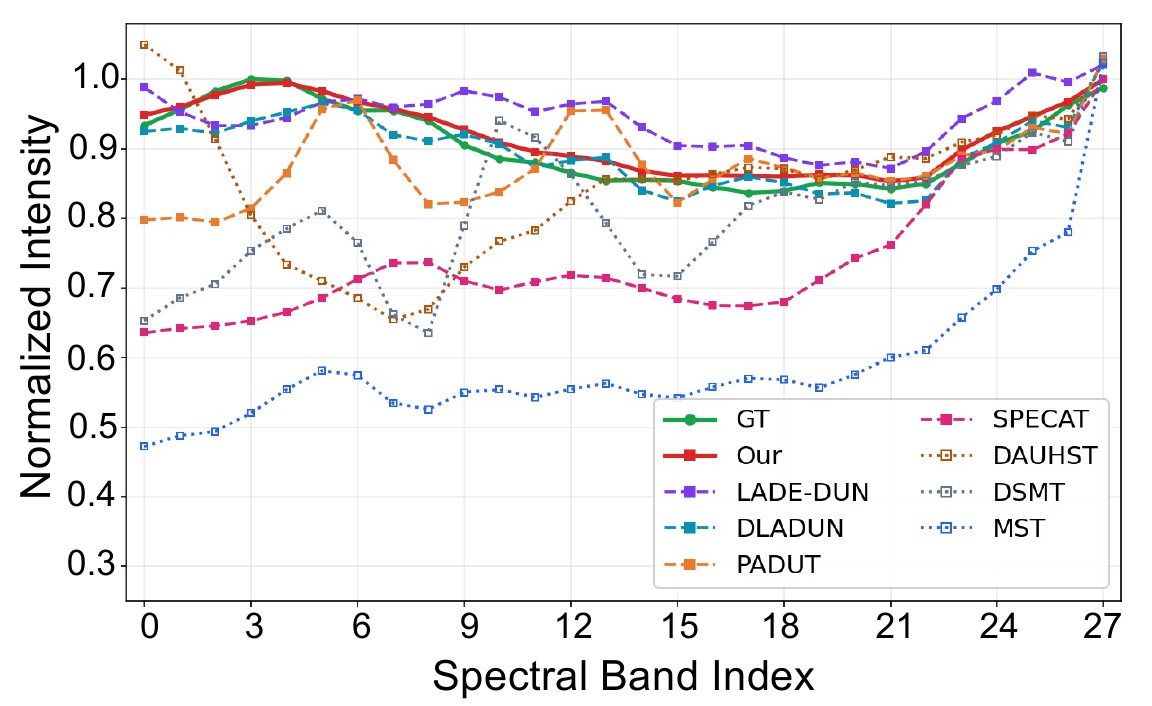}
    \caption{Spectral Intensity Profiles.}
    \label{fig1}
    \vspace{-6pt}
  \end{wrapfigure}

Hyperspectral Imaging (HSI) captures scenes across dense 
continuous spectral bands, enabling fine-grained material 
identification~\cite{li2020prior}, environmental 
monitoring~\cite{rajabi2024hyperspectral,elmasry2007hyperspectral}, precision medical imaging~\cite{lu2014medical}, and object tracking~\cite{li2022target,keshava2002spectral}. However, traditional scanning spectrometers rely on time-consuming line-by-line scanning, limiting their applicability to real-time applications. Based on Compressed Sensing (CS)~\cite{candes2006robust,zhang2016novel}, Snapshot 
Compressive Imaging (SCI) captures all spectral information in a 2D measurement, necessitating computational algorithms to recover the 3D HSI cube from highly underdetermined observations.

Recently, deep learning-based methods for HSI reconstruction can be broadly categorized into end-to-end learning methods~\cite{chen2021pre,meng2025sp3ctralmamba} and model-aided methods~\cite{li2023pixel,qiu2021effective,zheng2021deep}. End-to-end methods recover HSI via direct mapping but treat the imaging system as a black box. Model-aided methods, including Plug-and-Play (PnP)~\cite{qiu2021effective,zheng2021deep} and Deep Unfolding Networks (DUNs)~\cite{monga2021algorithm,gregor2010learning,li2023pixel}, leverage physical characteristics for inherent interpretability. Since PnP-based methods rely on fixed pretrained denoisers, their reconstruction performance remains limited. By contrast, deep unfolding-based methods have achieved significant improvements by incorporating learnable components into the iterative optimization process.

As the deep unfolding framework unfolds the iterative optimization algorithms with neural networks, it typically includes a projection module and a denoiser module. The data-consistency projection module relies solely on the sensing matrix, which encodes only the physical modulation properties of the imaging system. The resulting per-band scaling therefore reflects only system-determined modulation ratios rather than the true spectral energy distribution of the scene, causing the projection step to apply scene-agnostic spectral corrections that fail to account for varying energy contributions across bands, leading to inaccurate spectral energy allocation in the reconstructed HSI, as shown in Fig.~\ref{fig1}.
For the denoiser module, efficient feature extraction plays a critical role in the multi-stage optimization. However, existing denoisers either suffer from high computational complexity or limited receptive fields, failing to effectively capture global spatial-spectral dependencies.

To address these limitations, we propose a deep unfolding framework centered on manifold prior learning for HSI reconstruction. The manifold prior is a compact learned latent representation that captures the intrinsic spatial-spectral structure of clean HSIs, obtained through a two-stage strategy: Stage~1 trains a manifold compressor on clean HSIs to distill their intrinsic spatial-spectral structure into a compact latent code free from degradation interference, while Stage~2 employs a lightweight conditional diffusion model to infer this prior from compressed measurements. 
The inferred prior is then injected into both modules: in the data-consistency projection,a prior-guided spectral weight mechanism adaptively modulates the per-band contributions beyond the fixed system-determined scaling, correcting spectral energy allocation based on scene content; in the denoiser module, we introduce a manifold-guided Mamba that conditions the selective state-space scanning on the manifold prior, injecting global spatial-spectral structural context into every scan token to complement the local input-dependent adaptivity of Mamba's selective scanning.

In summary, our contributions can be  summarized as 
follows:

\noindent\textbf{(i)} We propose a deep unfolding 
framework with manifold prior injection into both the 
data-consistency projection and the denoiser module, ensuring that every unfolding stage receives degradation-free structural guidance rather than relying solely on corrupted intermediate estimates.

\noindent\textbf{(ii)} We introduce a manifold-guided Mamba denoiser that combines linear-complexity efficiency of SSM with global scene-level structural  context from the manifold prior, enabling structure-aware spatial-spectral feature extraction at each unfolding stage.

\noindent\textbf{(iii)} Quantitative and visual comparisons on datasets from both CASSI and optical filter-based imaging systems demonstrate that our method consistently outperforms existing state-of-the-art approaches while maintaining the physics-driven iterative structure.

\section{Related Work}

\subsection{Hyperspectral Imaging Systems}

Hyperspectral Imaging (HSI) has advanced from traditional scanning systems, including pushbroom~\cite{mouroulis2000design} and whiskbroom~\cite{zhang2016novel} systems, to high-efficiency snapshot imaging systems. Although conventional scanning methods provide high reconstruction accuracy, their inherent time delay makes them inapplicable for dynamic scenes and real-time applications. To address this, Snapshot Compressive Imaging (SCI) based on Compressed Sensing~\cite{candes2006robust} has been proposed to integrate high-dimensional spectral information into a single 2D sensor capture, necessitating computational algorithms to recover the 3D HSI cube from highly underdetermined observations.

Two representative paradigms have attracted significant attention: Coded Aperture Snapshot Spectral Imaging (CASSI)~\cite{gehm2007single,wagadarikar2009video} and optical filter-based architectures~\cite{xiong2022dynamic,yang2023deep}, which correspond to two physically different encoding mechanisms for spectral acquisition. CASSI systems rely on complex dispersive devices to encode spatial and spectral domains separately, resulting in bulky structures with poor mechanical stability and robustness~\cite{marois2023birefringent}. By contrast, optical filter-based architectures integrate broadband filters directly onto CMOS sensor arrays, realizing extreme miniaturization and high light throughput, but impose higher demands on the spatial-spectral decoupling ability of the reconstruction algorithm due to the high inter-band diversity of the sensing matrix.

\subsection{Hyperspectral Reconstruction Algorithms}

HSI reconstruction has evolved through four major paradigms. Model-based methods~\cite{wang2016adaptive,zhang2019computational} rely on hand-crafted priors such as total variation and sparsity, combined with iterative optimization solvers. They offer strong physical interpretability but lack representational capacity and require tedious manual tuning of regularization hyperparameters~\cite{dabov2007image,mairal2010online}, limiting their performance on complex real-world scenes. 
Plug-and-Play (PnP) methods~\cite{qiu2021effective,zheng2021deep,
lai2022deep,fu2022reinforcement} integrate pre-trained denoisers into iterative solvers as implicit regularizers, decoupling the data-fidelity and prior terms, but typically process spectral bands independently, neglecting intrinsic spatial-spectral correlations that are essential for high-fidelity HSI reconstruction. 
End-to-End (E2E) approaches directly learn the mapping from 2D measurements to 3D HSIs via CNNs~\cite{hu2022hdnet}, Transformers~\cite{cai2022cvpr,wang2022uformer,chen2021pre}, or Mamba-based architectures~\cite{meng2025sp3ctralmamba}; related E2E studies have further explored spatial-spectral concordance with unaligned RGB guidance~\cite{zhang2025unaligned}, abundance-level fusion~\cite{zhang2026enhancing}, degradation-aware prompting~\cite{wang2026metric}, and real-noise decoupling~\cite{zhang2026real} to improve HSI restoration. However, these methods generally treat the imaging system as a black box, ignoring the underlying imaging physics and spatial-spectral coupling.

Deep Unfolding Networks (DUNs)~\cite{monga2021algorithm,gregor2010learning,wang2020dnu,li2023pixel,li2025interferometric} unroll iterative optimization steps into a fixed-depth learnable network, combining physical interpretability with the representational power of deep learning, and have achieved superior reconstruction by alternating between a data-consistency projection step and a learned denoiser at each stage. More recently, some DUNs have incorporated latent diffusion models~\cite{rombach2022high,wu2024} to generate degradation-free priors for the denoiser module, further improving reconstruction quality. Diffusion-based priors have also been explored for multimodal PET restoration, demonstrating their broader potential for degradation-aware image restoration~\cite{zhang2026supervise}. Despite these advances, the projection step remains constrained by fixed system-determined spectral scaling that ignores scene content, while the denoiser operates on degradation-contaminated inputs without clean structural references, and the prior is injected only into the denoiser, leaving the projection step without scene-adaptive guidance. Furthermore, Transformer-based denoisers incur quadratic complexity that limits scalability to high-resolution HSI cubes. In this work, we address these limitations by proposing a manifold prior guided deep unfolding framework that injects degradation-free structural priors into both the projection and denoiser modules, and adopts a Mamba-based denoiser for efficient linear-complexity spatial-spectral feature extraction.

\section{Proposed Method}

\subsection{Problem Formulation}
HSI reconstruction can be formulated as a standard linear inverse problem:
\begin{equation}
    \mathbf{y} = \mathbf{A}\mathbf{x} + \mathbf{b},
\end{equation}
where $\mathbf{A}$ encodes the imaging physics, and $\mathbf{b}$ is noise. Recovering $\mathbf{x}$ from the highly underdetermined $\mathbf{y}$ requires effective priors to regularize the solution space~\cite{ongie2020deep}. 
We cast HSI reconstruction as a constrained optimization problem and unfold the iterative algorithm into a learnable deep network. The standard regularized inverse problem is:
\begin{equation}
    \hat{\mathbf{x}} = \arg\min_{\mathbf{x}} \frac{1}{2}\|\mathbf{y}-\mathbf{A}\mathbf{x}\|_2^2 + \tau\mathcal{R}(\mathbf{x}),
\end{equation}
where $\tau$ balances data fidelity and the image prior $\mathcal{R}(\mathbf{x})$. We employ Generalized Alternating Projection (GAP)~\cite{yuan2016generalized} as the optimization framework, and by introducing an auxiliary variable $\mathbf{z}$, the problem is reformulated as:
\begin{equation}
    (\hat{\mathbf{x}}, \hat{\mathbf{z}}) = \arg\min_{\mathbf{x},\mathbf{z}} \frac{1}{2}\|\mathbf{x}-\mathbf{z}\|_2^2 + \lambda\mathcal{R}(\mathbf{z}), \quad \text{s.t.} \quad \mathbf{y}=\mathbf{A}\mathbf{x},
\end{equation}
which decomposes into two alternating sub-problems: the projection step enforces data consistency via Euclidean projection onto the measurement subspace:
\begin{equation}
    \mathbf{x}^{(k+1)} = \mathbf{z}^{(k)} + \mathbf{A}^{\top}(\mathbf{A}\mathbf{A}^{\top})^{-1}(\mathbf{y} - \mathbf{A}\mathbf{z}^{(k)}),
\end{equation}
and the proximal step incorporates the image prior via a learnable denoiser:
\begin{equation}
    \mathbf{z}^{(k+1)} = \mathcal{D}_{k}(\mathbf{x}^{(k+1)}),
\end{equation}
where $\mathcal{D}_{k}(\cdot)$ denotes the learnable denoiser at the $k$-th stage.

For the CASSI and optical filter-based systems, $\mathbf{A}\mathbf{A}^{\top}$ is diagonal, so its exact inverse reduces to element-wise division by $\mathbf{\Phi}_s = \mathrm{diag}(\mathbf{A}\mathbf{A}^{\top})$, yielding the unfolded projection:
\begin{equation}
    \mathbf{x}^{(k+1)} = \mathbf{z}^{(k)} + \mathbf{A}^{\top}(\mathbf{y} - \mathbf{A}\mathbf{z}^{(k)}) / \mathbf{\Phi}_s.
    \label{eq:projection_unfolded}
\end{equation}

However, $\mathbf{\Phi}_s$ is fully determined by the physical structure of the sensing matrix and carries no information about the spectral energy distribution of the scene. The correction applied to each spectral band is therefore governed solely by system-determined modulation ratios, regardless of the actual energy contributions of individual bands, providing the denoiser with a spectrally biased estimate at each stage.

\subsection{Manifold Prior Guided Deep Unfolding Framework}
The above analysis reveals two coupled limitations in conventional DUNs: scene-agnostic per-band corrections dictated solely by $\mathbf{\Phi}_s$, and denoising on degradation-contaminated estimates without clean structural reference. To address both simultaneously, we propose to distill the intrinsic spatial-spectral structure of clean HSIs into a compact manifold prior $\mathbf{p}$, and strategically inject it into both the data-consistency projection via a Prior-Guided Spectral Weight mechanism and the denoiser via a Manifold-Guided Mamba, providing degradation-free structural guidance at every unfolding stage.

Our manifold prior $\mathbf{p}$ is not a generic conditioning signal, but an empirical coordinate representation on a low-dimensional submanifold. Here, we provide its formal mathematical formulation and physical realization.
Following the Low-Dimensional Manifold Model (LDMM)~\cite{zhu2020ldmm}, although a spatial-spectral patch extracted from a hyperspectral image $\mathbf{z}$ has a high ambient dimension after vectorization, its variations are governed by only a few intrinsic factors. Specifically, for an arbitrary spatial location $x\in\bar{\Omega}$, a patch of size $s_1 \times s_2 \times B$ is vectorized as:
\begin{equation}
    \mathbf{q}_x
    =
    \operatorname{vec}\bigl(\mathcal{E}_{\mathrm{patch}}(\mathbf{z},x)\bigr)
    \in\mathbb{R}^{D},
    \qquad
    D=s_1s_2B.
\end{equation}
where $\mathcal{E}_{\mathrm{patch}}$ denotes the patch-extraction operator. All patches constitute a point cloud:
\begin{equation}
    \mathcal{Q}(\mathbf{z})
    =
    \left\{
        \mathbf{q}_x:x\in\bar{\Omega}
    \right\}
    \subset\mathbb{R}^{D}.
\end{equation}
This point cloud is assumed to lie on or close to a union of low-dimensional smooth submanifolds:
\begin{equation}
    \mathcal{M}
    =
    \bigcup_{\ell=1}^{L}\mathcal{M}_{\ell}
    \subset\mathbb{R}^{D}.
    \label{eq:manifold_union}
\end{equation}
From a theoretical parameterization perspective, each manifold branch can be expressed as:
\begin{equation}
    \mathcal{M}_{\ell}
    =
    \left\{
        g_{\ell}(\mathbf{p}_x) :
        \mathbf{p}_x\in\mathcal{Z}_{\ell}
    \right\},
    \label{eq:manifold_parameterization}
\end{equation}
where $\mathcal{Z}_{\ell}\subset\mathbb{R}^{k}$ denotes the low-dimensional intrinsic coordinate domain, $\mathbf{p}_x$ denotes the intrinsic coordinate of a patch $\mathbf{q}_x$, and $g_{\ell}(\cdot)$ maps the coordinate to the corresponding high-dimensional patch representation. Let $\mathcal{Z}=\bigcup_{\ell=1}^{L}\mathcal{Z}_{\ell}$ denote the global latent coordinate space.

As formalized above, while the vectorized patch $\mathbf{q}_x$ resides in a high-dimensional space $\mathbb{R}^{D}$, its intrinsic geometric state is approximately parameterized by the low-dimensional coordinate $\mathbf{p}_x\in\mathcal{Z}$. LDMM~\cite{zhu2020ldmm} circumvents the unknown analytical form of $g_{\ell}(\cdot)$ by constructing a graph Laplacian $\mathbf{L}_{\mathrm{graph}}$ over patch distances to approximate manifold operators.

In contrast, rather than performing computationally heavy graph construction, we propose to directly estimate the global manifold prior $\mathbf{p}$. Specifically, our Manifold Compressor $\mathcal{E}_{\mathrm{mc}}$ acts as a data-driven encoder, mapping the high-dimensional HSI data $\mathbf{z}$ onto the low-dimensional domain to yield the empirical manifold coordinate:
\begin{equation}
    \mathbf{p}
    =
    \mathcal{E}_{\mathrm{mc}}(\mathbf{z})
    \in\mathcal{Z}.
    \label{eq:manifold_prior}
\end{equation}
This learned global coordinate $\mathbf{p}$ encapsulates the essential long-range spatial-spectral dependencies and directly conditions the subsequent unfolding stages.

\begin{figure}[t]
    \centering
    \includegraphics[width=1\linewidth]{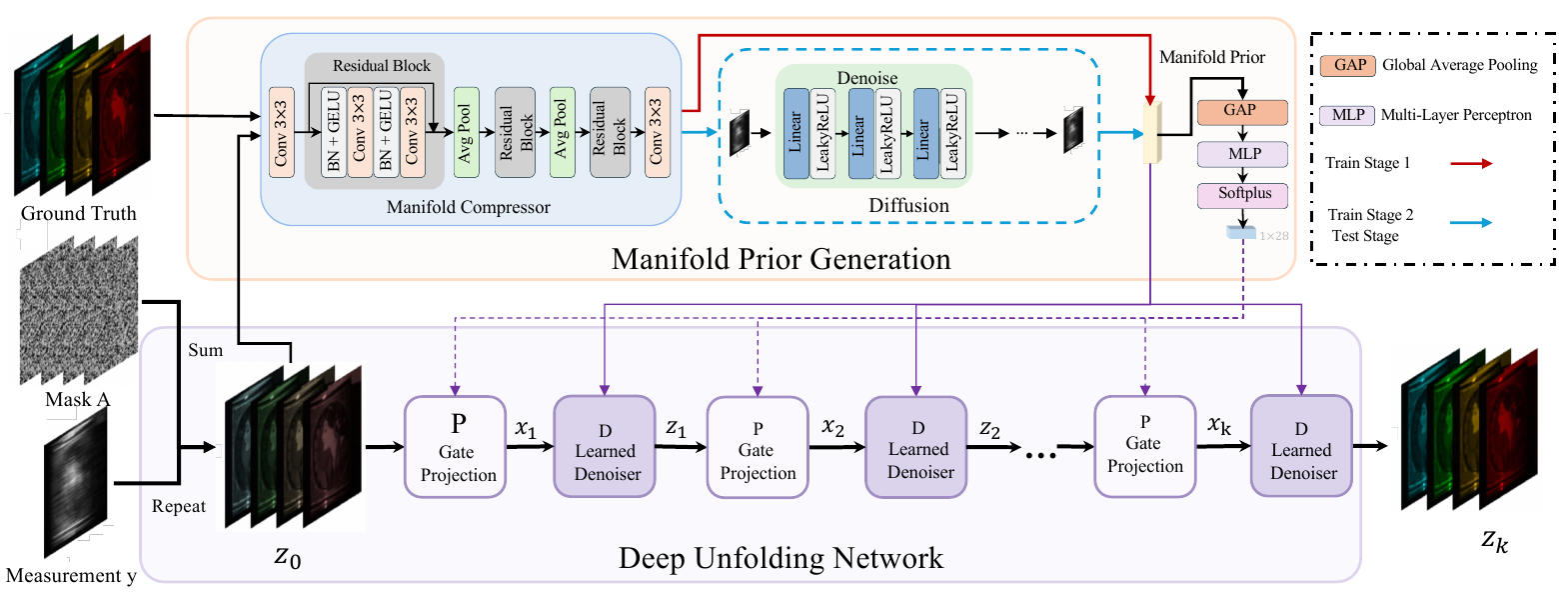}
      \caption{Proposed prior-guided deep unfolding framework for hyperspectral image reconstruction, 
      where a learned manifold prior guides iterative projection and denoiser from $\mathbf{z}_0$ to $\mathbf{z}_k$.}
    \label{fig2}
\end{figure}

\subsection{Manifold Prior Generation and Guided Spectral Weight}
As Fig.~\ref{fig2} shows, we learn a manifold prior 
$\mathbf{p}$ (or its estimate $\hat{\mathbf{p}}$) from 
data and use it to adaptively guide reconstruction through a coordinated two-stage process.

\noindent\textbf{Two-Stage Manifold Prior Generation.} Although the manifold prior can be learned from clean HSI data during training, the clean reference is unavailable at inference time. To bridge this gap, we adopt a two-stage strategy: Stage~1 learns a reliable manifold prior from the clean reference and the back-projected initialization, while Stage~2 trains a conditional diffusion model to infer the corresponding prior from the compressed measurement alone. This design enables the learned structural knowledge to be effectively transferred from training to inference and used for prior-guided unfolding reconstruction.

\noindent\textit{Stage 1 (Manifold Prior Learning):}
Although clean HSIs are nominally high-dimensional, their spatial-spectral redundancy suggests the existence of a compact low-dimensional representation. We exploit this structure by training a Manifold Compressor $\mathcal{E}_{\mathrm{mc}}$ to map clean HSIs onto a low-dimensional manifold coordinate $\mathbf{p}$ that encodes scene-specific structural and spectral characteristics. $\mathcal{E}_{\mathrm{mc}}$ is implemented as a hierarchical convolutional encoder with residual blocks and progressive downsampling, which progressively discards redundant spatial-spectral information while retaining the essential manifold structure of the scene. Given the back-projected initialization $\mathbf{z}_0 \in \mathbb{R}^{H \times W \times N_\lambda}$ and the clean reference $\mathbf{z}^* \in \mathbb{R}^{H \times W \times N_\lambda}$, $\mathcal{E}_{\mathrm{mc}}$ takes their concatenation as input and produces the manifold prior:
\begin{equation}
    \mathbf{p} = \mathcal{E}_{\mathrm{mc}}([\mathbf{z}^*;\,\mathbf{z}_0]),
\end{equation}
where $[\cdot;\cdot]$ denotes channel-wise concatenation. The progressive downsampling in $\mathcal{E}_{\mathrm{mc}}$ acts as a geometric projection onto the low-dimensional manifold, filtering out degradation-irrelevant details and retaining only the structural and spectral essence of the clean HSI.
Stage~1 is optimized with:
\begin{equation}
    \mathcal{L}_1 = \|\mathbf{z}^*-\mathbf{z}_k\|_{\mathrm{CB}},
\end{equation}
where $\|\cdot\|_{\mathrm{CB}}$ denotes the Charbonnier loss. Through end-to-end training with the reconstruction objective, $\mathcal{E}_{\mathrm{mc}}$ learns to extract precisely the manifold coordinates that are most informative for guiding high-quality HSI reconstruction.

\noindent\textit{Stage 2 (Diffusion-Based Prior Inference):}
In this stage, $\mathcal{E}_{\mathrm{mc}}$ provides ground-truth manifold prior samples $\mathbf{p}$ as supervision targets. We train a conditional diffusion model to infer $\hat{\mathbf{p}}$ from $\mathbf{z}_0$ alone, without access to the clean reference. The forward diffusion process synthesizes a noisy prior:
\begin{equation}
    \mathbf{p}_t
    = \sqrt{\bar{\alpha}_t}\,\mathbf{p}
    + \sqrt{1-\bar{\alpha}_t}\,\boldsymbol{\epsilon},
    \quad \boldsymbol{\epsilon}\sim\mathcal{N}(\mathbf{0},\mathbf{I}),
\end{equation}
where $\bar{\alpha}_t=\prod_{i=1}^{t}(1-\beta_i)$. The reverse process is conditioned on $\mathbf{z}_0$ via the Manifold Compressor $\mathcal{E}_{\mathrm{mc}}$:
\begin{equation}
    \mathbf{c} = \mathcal{E}_{\mathrm{mc}}(\mathbf{z}_0).
\end{equation}
At each reverse step, the denoiser module $\mathcal{D}_\theta$ takes the noisy prior, condition, and normalized timestep as input:
\begin{equation}
    \hat{\mathbf{p}}
    = \mathcal{D}_\theta\!\left(
        [\mathbf{p}_t;\,\mathbf{c};\,\tilde{t}]
    \right),
\end{equation}
where $\tilde{t}=t/T\in[0,1]$ is the normalized timestep. The inferred $\hat{\mathbf{p}}$ serves directly as the manifold prior to guide the subsequent reconstruction. Stage~2 is optimized via:
\begin{equation}
    \mathcal{L}_2
    = \|\mathbf{z}^*-\mathbf{z}_k\|_{\mathrm{CB}}
    + \|\mathbf{p}-\hat{\mathbf{p}}\|_{\mathrm{CB}},
\end{equation}
where $\mathbf{p}$ from Stage~1 serves as the supervision target.

\noindent\textbf{Prior-Guided Spectral Weight.}\quad 
We leverage the manifold prior to dynamically adjust the spectral filtering strength in the data-consistency module. Specifically, we generate spectral-wise weights from $\mathbf{p}$ or $\hat{\mathbf{p}}$ via a lightweight MLP:
\begin{equation}
    \mathbf{w}_\text{spec} = \mathrm{Softplus}\left(
    \mathrm{MLP}\left(\mathrm{Pool}(\mathbf{p})\right) + \mathbf{b}_0
    \right),
\end{equation}
where $\mathrm{Pool}(\cdot)$ performs global pooling over spatial dimensions to obtain a compact representation, the Softplus activation guarantees $\mathbf{w}_\text{spec}> 0$ to preserve physical consistency, $\mathbf{b}_0$ is the bias, and $\mathbf{w}_\text{spec}\in\mathbb{R}^{N_\lambda}$ contains one weight per spectral band. 

These weights dynamically rescale the per-band correction term in Eq.(6), 
allowing the model to adaptively modulate each spectral band according to the scene-specific 
spectral energy distribution inferred from the manifold prior:
\begin{equation}
    \mathbf{x}^{(k+1)} = \mathbf{z}^{(k)} + \left(\mathbf{A}^{\top}(\mathbf{y} - 
    \mathbf{A}\mathbf{z}^{(k)}) / \mathbf{\Phi}_s\right) \odot \mathbf{w}_{\text{spec}},
    \label{eq:projection_guided}
\end{equation}

\subsection{Mamba-Based Denoiser with Prior Modulation}

The denoiser module must efficiently capture both global context and fine spatial-spectral details. While Transformer-based denoisers excel at global modeling, their quadratic complexity $\mathcal{O}(N^2)$ limits scalability. We instead employ the Mamba architecture~\cite{gu2023mamba} based on Selective State Space Models (SSMs), which achieves linear complexity $\mathcal{O}(N)$ by dynamically generating parameters from the input at each position, enabling content-aware selective retention of information. To extend Mamba to 2D images, we adopt the 2D Selective Scanning (SS2D) mechanism~\cite{liu2024vmamba,zhu2024vision}, which flattens a 2D feature map into four directional sequences and processes each independently:
\begin{equation}
    \mathbf{f}^d = \mathrm{SS2D}^d(\mathbf{x}^d), \quad d \in \{1,2,3,4\}.
\end{equation}
The four outputs are summed and reshaped to recover global 2D perception:
\begin{equation}
    \mathbf{F}_{\mathrm{SS2D}} = \mathrm{Reshape}\!\left(\sum_{d=1}^4 \mathbf{f}^d\right).
\end{equation}

\noindent\textbf{Prior-Guided Mamba.}\quad 
While Mamba adapts its parameters to local input features, it lacks access to global scene-level structural information, as the scanning process can only accumulate context sequentially along fixed directions. To address this, we inject the manifold prior to modulate the SS2D input features $\mathbf{x}_k$ before scanning, providing each spatial location with a globally-informed context and steering the scanning process toward scene-consistent spatial-spectral representations.

\begin{figure*}[t]
  \centering
  \includegraphics[width=1\textwidth]{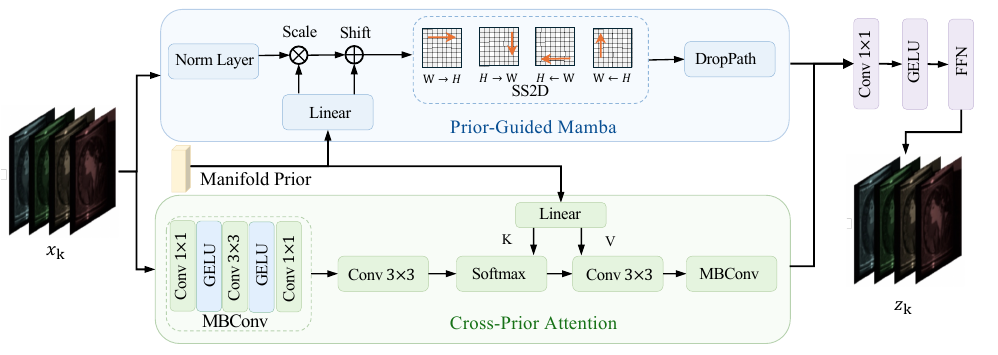}
    \caption{Proposed prior-guided denoiser module, combining a prior-modulated Mamba branch and a cross-prior attention branch for structure-aware feature refinement.}
  \label{fig3}
\end{figure*}

As illustrated in the Prior-Guided Mamba branch (Fig.~\ref{fig3}), the manifold prior $\mathbf{p}$ or $\hat{\mathbf{p}}$ is first projected via a linear layer and spatially interpolated to generate per-pixel modulation parameters:
\begin{equation}
    \mathbf{s}, \mathbf{t} = \mathrm{split}\left(
    \mathrm{Interp}\left(\mathrm{Linear}_{\mathrm{prior}}
    (\mathbf{p})\right)\right),
\end{equation}
where $\mathrm{Interp}(\cdot)$ denotes bilinear interpolation to match the spatial resolution of the feature map, and the output is evenly split into a scale factor $\mathbf{s}$ and a shift factor $\mathbf{t}$. 
The input feature $\mathbf{x}_k$ is then normalized and modulated as:
\begin{equation}
    \tilde{\mathbf{x}}_k = \mathrm{Norm}(\mathbf{x}_k) \odot (1 + \mathbf{s}) + \mathbf{t},
\end{equation}
where $\odot$ denotes element-wise multiplication and $\mathrm{Norm}(\cdot)$ denotes layer normalization. The scale factor $\mathbf{s}$ and shift factor $\mathbf{t}$ adaptively recalibrate the feature distribution at each spatial location based on the global scene context encoded in the manifold prior.

The modulated features $\tilde{\mathbf{x}}_k$ are flattened along four scanning directions ($d \in \{1,2,3,4\}$) and fed into the SS2D module. This prior-guided modulation complements the local input-dependent adaptation  with  global scene-level context, steering the scanning process toward scene-consistent spatial-spectral representations and enabling feature extraction aligned with the genuine HSI geometry.
 
\noindent\textbf{Cross-Prior Attention and Overall Denoiser.}\quad 
In parallel to the Prior-Guided Mamba branch, we introduce a Cross-Prior Attention module to further exploit the structural information encoded in $\mathbf{p}$ or $\hat{\mathbf{p}}$. 
Specifically, the input feature $\mathbf{x}_k$ is first processed by a MBConv block for channel mixing and local feature extraction, followed by a $3\times3$ convolution to further capture spatial context:
\begin{equation}
    \mathbf{Q} = \mathrm{Conv}_{3\times3}\big(\mathrm{MBConv}(\mathbf{x})\big).
\end{equation}
The manifold prior $\mathbf{p}$ or $\hat{\mathbf{p}}$ is first projected via a linear layer to generate key and value embeddings:
\begin{equation}
    [\mathbf{K}, \mathbf{V}] = \mathrm{Linear}(\mathbf{p}).
\end{equation}
The interaction between the feature and the prior is performed through a cross-feature aggregation mechanism, where the query $\mathbf{Q}$ interacts with prior-derived representations to selectively incorporate structurally relevant information from the manifold prior. This process enhances content-aware feature responses by injecting global structural cues into local feature representations. The aggregated features are subsequently refined through a $3\times3$ convolution to improve local spatial consistency, followed by an MBConv block for efficient channel mixing and nonlinear transformation.

The outputs from the Prior-Guided Mamba branch and the Cross-Prior Attention branch are concatenated and fused through a $1\times1$ convolution followed by GELU activation. A residual connection and a feed-forward network (FFN) are further applied to produce the final output feature $\mathbf{z}_k$.

\begin{table*}[htbp]
\centering
\caption{Comparisons between our method and SOTA methods on 10 simulation scenes (S1$\sim$S10) on the simulated Optical Filter-based HSI system. Params, FLOPs, PSNR (upper entry in each cell), and SSIM (lower entry in each cell) are reported.}
\label{tab1}
\renewcommand{\arraystretch}{1.15} 
\setlength{\heavyrulewidth}{1.5pt} 
\setlength{\tabcolsep}{4.2pt}       
\resizebox{\textwidth}{!}{%
\begin{tabular}{lcc ccccccccccc}
\toprule 
Method & Params & FLOPs & S1 & S2 & S3 & S4 & S5 & S6 & S7 & S8 & S9 & S10 & Avg \\
\midrule 

\multirow{2}{*}{MST \cite{cai2022cvpr}} & \multirow{2}{*}{2.03M} & \multirow{2}{*}{28.15G} 
& 36.55 & 36.29 & 33.46 & 39.78 & 35.40 & 36.10 & 34.53 & 33.08 & 34.38 & 35.19 & 35.48 \\
& & & 0.963 & 0.953 & 0.904 & 0.951 & 0.964 & 0.963 & 0.924 & 0.945 & 0.926 & 0.966 & 0.946 \\
\midrule

\multirow{2}{*}{DAUHST-9stg \cite{cai2022neurips}} & \multirow{2}{*}{6.15M} & \multirow{2}{*}{79.50G} 
& 38.37 & 39.91 & 37.71 & 42.97 & 37.69 & 39.05 & 37.62 & 36.11 & 38.45 & 37.39 & 38.53 \\
& & & 0.972 & 0.977 & 0.965 & 0.967 & 0.980 & 0.974 & 0.964 & 0.965 & 0.971 & 0.976 & 0.971 \\
\midrule

\multirow{2}{*}{PADUT-12stg \cite{li2023pixel}} & \multirow{2}{*}{5.38M} & \multirow{2}{*}{90.46G} 
& 38.42 & 40.34 & 38.95 & 43.50 & 38.22 & 39.16 & 38.21 & 36.03 & 39.45 & 37.30 & 38.89 \\
& & & 0.974 & 0.983 & 0.972 & 0.977 & 0.983 & 0.979 & 0.971 & 0.969 & 0.979 & 0.981 & 0.977 \\
\midrule

\multirow{2}{*}{SPECAT \cite{yao2024}} & \multirow{2}{*}{0.29M} & \multirow{2}{*}{12.40G} 
& \underline{40.24} & 42.40 & 41.43 & 44.90 & \underline{39.62} & \underline{39.90} & 39.41 & 37.49 & 40.45 & \underline{37.90} & 40.37 \\
& & & 0.982 & 0.986 & 0.978 & 0.982 & 0.987 & 0.984 & 0.977 & 0.977 & 0.982 & 0.983 & 0.986 \\
\midrule

\multirow{2}{*}{LADE-DUN-3stg \cite{wu2024}} & \multirow{2}{*}{2.56M} & \multirow{2}{*}{32.92G} 
& 37.08 & 39.53 & 41.67 & 46.37 & 35.73 & 36.71 & 36.94 & 35.13 & 39.96 & 33.96 & 38.31 \\
& & & 0.961 & 0.974 & 0.978 & 0.991 & 0.972 & 0.973 & 0.962 & 0.971 & 0.976 & 0.961 & 0.972 \\
\midrule

\multirow{2}{*}{LADE-DUN-9stg \cite{wu2024}} & \multirow{2}{*}{3.06M} & \multirow{2}{*}{86.23G} 
& 39.72 & \underline{42.87} & \underline{41.36} & \textbf{47.15} & 39.36 & \textbf{40.03} & \underline{39.54} & \underline{37.32} & \underline{40.81} & 37.54 & \underline{40.57} \\
& & & 0.982 & \underline{0.988} & 0.982 & \underline{0.992} & 0.988 & 0.988 & 0.977 & \underline{0.981} & \underline{0.986} & 0.981 & \underline{0.984} \\
\midrule

\multirow{2}{*}{DSMT \cite{luo2025dsmt}} & \multirow{2}{*}{12.40M} & \multirow{2}{*}{44.94G} 
& 36.85 & 38.10 & 39.45 & 43.20 & 34.60 & 36.90 & 35.80 & 34.90 & 38.15 & 34.10 & 37.21 \\
& & & 0.961 & 0.968 & 0.970 & 0.983 & 0.962 & 0.971 & 0.958 & 0.966 & 0.970 & 0.959 & 0.967 \\
\midrule

\multirow{2}{*}{DLADUN-9stg \cite{liu2026degradation}} & \multirow{2}{*}{3.94M} & \multirow{2}{*}{196.26G} 
& 39.02 & 41.55 & \underline{41.83} & 46.51 & 38.92 & 39.75 & 38.82 & \underline{38.05} & \textbf{42.10} & 36.25 & 40.28 \\
& & & 0.976 & 0.986 & 0.982 & 0.992 & 0.985 & 0.985 & 0.972 & \underline{0.984} & 0.985 & 0.974 & 0.982 \\
\midrule

\multirow{2}{*}{\textbf{Ours-3stg}} & \multirow{2}{*}{3.71M} & \multirow{2}{*}{33.62G} 
& 37.45 & 40.01 & 42.12 & 46.80 & 36.20 & 37.15 & 37.40 & 35.60 & 40.40 & 34.45 & 38.76 \\
& & & 0.965 & 0.978 & 0.982 & 0.993 & 0.976 & 0.977 & 0.968 & 0.974 & 0.981 & 0.966 & 0.976 \\
\midrule

\multirow{2}{*}{\textbf{Ours-9stg}} & \multirow{2}{*}{3.71M} & \multirow{2}{*}{70.62G} 
& \textbf{40.55} & \textbf{43.70} & \textbf{42.95} & \underline{46.85} & \textbf{40.10} & \textbf{40.42} & \textbf{40.25} & \textbf{38.20} & \underline{42.05} & \textbf{38.65} & \textbf{41.37} \\
& & & \textbf{0.986} & \textbf{0.992} & \textbf{0.986} & \textbf{0.995} & \textbf{0.991} & \textbf{0.991} & \textbf{0.982} & \textbf{0.985} & \textbf{0.991} & \textbf{0.991} & \textbf{0.990} \\
\bottomrule 
\end{tabular}}
\end{table*}

\section{Experiments}

\subsection{Experiment Setup}

\textbf{Optical Filter-based HSI System Data.}
We utilize two benchmark datasets: CAVE~\cite{park2007iccv} and KAIST~\cite{choi2017kaist}, both covering 28 spectral channels ranging from 450\,nm to 650\,nm. The CAVE dataset contains 32 HSIs with a spatial resolution of $512\times512$ pixels, while KAIST contains 30 HSIs at $2704\times3376$ pixels. Following Yao et al.~\cite{yao2024}, all data are additionally processed with an optical filter-based mask, adopting the Fabry-P\'{e}rot design of Yako et al.~\cite{yako2023} as a representative optical filter-based HSI system.

\textbf{CASSI System Data.}
Following~\cite{meng2020eccv,hu2022cvpr,huang2021cvpr}, CAVE is used for training and 10 scenes from KAIST for testing, with simulated measurements generated via a CASSI system mask. Furthermore, we evaluate on real HSI measurements captured by the CASSI system established in TSA-Net~\cite{meng2020eccv}. To better reflect authentic sensor behavior, noise is incorporated into the training measurements to approximate real-world projection acquisition, following the realistic-noise setting commonly considered in HSI restoration~\cite{zhang2021hyperspectral}.

\textbf{Metrics and Implementation Specifics.} Reconstruction performance is evaluated using PSNR and SSIM~\cite{wang2004image}. The model is implemented in PyTorch~\cite{paszke2019pytorch} and trained on a single NVIDIA RTX 3090 GPU using Adam with an initial learning rate of $4\times10^{-4}$ decayed via cosine annealing to $1\times10^{-6}$ over 100 epochs. Training follows a two-phase strategy: Phase~1 (epochs 1--30) optimizes the reconstruction backbone with $\mathcal{L}_1$; Phase~2 (epochs 31--100) introduces the diffusion prior and optimizes with $\mathcal{L}_2$, which adds a manifold prior alignment loss to supervise the inferred prior.

\subsection{Results and Discussion}
To provide a comprehensive evaluation, we compare our method against seven state-of-the-art approaches spanning major methodological paradigms in HSI reconstruction. 
These include the Transformer-based methods MST~\cite{cai2022cvpr} and SPECAT~\cite{yao2024}, which exploit long-range spatial-spectral modeling; 
the deep unfolding methods PADUT~\cite{li2023pixel}, DLADUN~\cite{liu2026degradation}, and DAUHST~\cite{cai2022neurips}, which incorporate physics-driven iterative optimization;
the recent diffusion-prior deep unfolding method LADE-DUN~\cite{wu2024}; 
and the hybrid method DSMT~\cite{luo2025dsmt}, which combines spectral and spatial modeling. 
All comparisons are conducted under identical experimental settings to ensure fairness.

\noindent\textbf{Results of Optical Filter-based HSI System.}
Tab.~\ref{tab1} reports the quantitative results on the optical filter-based HSI system simulation dataset. 
Ours-9stg achieves state-of-the-art performance with 41.37\,dB in PSNR and 0.990 in SSIM, outperforming the recent competitive framework LADE-DUN-9stg~\cite{wu2024} by 0.80\,dB and the deep unfolding model DAUHST-9stg~\cite{cai2022neurips} by 2.84\,dB. 
Notably, compared with DAUHST-9stg~\cite{cai2022neurips}, our model achieves 2.84\,dB higher PSNR while requiring only 60.3\% of the parameters and 88.8\% of the FLOPs. 
Against the Transformer-based baseline MST~\cite{cai2022cvpr}, our method gains a substantial 5.89\,dB. 
Even with a limited number of unfolding iterations, Ours-3stg achieves a PSNR improvement of 0.45\,dB over the corresponding 3-stage LADE-DUN. 
These results demonstrate the effectiveness of the proposed manifold prior guidance for optical filter-based HSI reconstruction.

\textbf{Results of CASSI System.} Tab.~\ref{tab2} presents the quantitative results on the CASSI system simulation dataset. Ours-9stg achieves 40.24\,dB in PSNR and 0.980 in SSIM, outperforming all competing methods on average. Specifically, Ours-9stg surpasses SPECAT~\cite{yao2024} by 1.67\,dB, LADE-DUN-9stg~\cite{wu2024} by 0.33\,dB, and DAUHST-9stg~\cite{cai2022neurips} by 1.88\,dB. Even with a limited number of stages, our lightweight Ours-3stg achieves 38.73\,dB, which represents a 0.42\,dB improvement over the 3-stage LADE-DUN (38.31\,dB) and also exceeds the performance of the Transformer-based SPECAT (38.57\,dB). These consistent gains confirm that the proposed manifold prior guidance generalizes effectively to the dispersive modulation of CASSI systems.

\newcommand{\sd}[1]{\scalebox{0.5}{#1}}
\newcommand{\sa}[1]{\scalebox{0.7}{#1}}
\begin{table*}[htbp]
\centering
\caption{Comparisons between our method and SOTA methods on 10 simulation scenes (S1$\sim$S10) on the simulated CASSI system. PSNR (left) / SSIM (right) are reported in each cell.}
\label{tab2}
\renewcommand{\arraystretch}{1.2} 
\setlength{\tabcolsep}{3pt}
\begin{tabular}{l cccccccccccc}
\Xhline{1.0pt}
\sa{Method} & \sa{S1} & \sa{S2} & \sa{S3} & \sa{S4} & \sa{S5} & \sa{S6} & \sa{S7} & \sa{S8} & \sa{S9} & \sa{S10} & \sa{Avg} \\
\Xhline{0.8pt}
\sa{MST~\cite{cai2022cvpr}} & \sd{35.40/0.941} & \sd{35.87/0.944} & \sd{36.51/0.953} & \sd{42.27/0.973} & \sd{32.77/0.947} & \sd{34.80/0.955} & \sd{33.66/0.925} & \sd{32.67/0.948} & \sd{35.39/0.949} & \sd{32.50/0.941} & \sd{35.18/0.948} \\
\hline
\sa{DAUHST-9stg~\cite{cai2022neurips}} & \sd{37.25/0.958} & \sd{39.02/0.967} & \sd{41.05/0.971} & \sd{46.15/0.983} & \sd{35.80/0.969} & \sd{37.08/0.970} & \sd{37.57/0.963} & \sd{35.10/0.966} & \sd{40.02/0.970} & \sd{34.59/0.956} & \sd{38.36/0.967} \\
\hline
\sa{PADUT-12stg~\cite{li2023pixel}} & \sd{37.36/0.962} & \sd{40.43/0.978} & \sd{42.38/0.979} & \sd{46.62/0.990} & \sd{36.26/0.974} & \sd{37.27/0.974} & \sd{37.83/0.966} & \sd{35.33/0.974} & \sd{40.86/0.978} & \sd{34.55/0.963} & \sd{38.89/0.974} \\
\hline
\sa{SPECAT~\cite{yao2024}} & \sd{\underline{38.12}/\textbf{0.977}} & \sd{40.79/\textbf{0.984}} & \sd{38.59/0.973} & \sd{43.39/0.985} & \sd{\underline{38.02}/\textbf{0.984}} & \sd{\textbf{38.41}/\textbf{0.984}} & \sd{37.82/0.965} & \sd{34.80/0.971} & \sd{38.75/0.980} & \sd{\textbf{37.04}/\textbf{0.986}} & \sd{38.57/0.979} \\
\hline
\sa{LADE-DUN-3stg~\cite{wu2024}} & \sd{37.08/0.961} & \sd{39.53/0.974} & \sd{41.67/0.978} & \sd{46.37/0.991} & \sd{35.73/0.972} & \sd{36.71/0.973} & \sd{36.94/0.962} & \sd{35.13/0.971} & \sd{39.96/0.976} & \sd{33.96/0.961} & \sd{38.31/0.972} \\
\hline
\sa{LADE-DUN-9stg~\cite{wu2024}} & \sd{38.07/0.969} & \sd{41.16/0.982} & \sd{\underline{43.70}/\textbf{0.983}} & \sd{\textbf{48.01}/\textbf{0.993}} & \sd{37.76/0.980} & \sd{37.65/\underline{0.980}} & \sd{\underline{38.58}/\underline{0.973}} & \sd{\underline{36.31}/\textbf{0.979}} & \sd{\underline{42.66}/\underline{0.984}} & \sd{\underline{35.18}/0.967} & \sd{\underline{39.91}/\underline{0.979}} \\
\hline
\sa{DSMT~\cite{luo2025dsmt}} & \sd{36.38/0.958} & \sd{37.86/0.965} & \sd{39.18/0.968} & \sd{43.89/0.985} & \sd{34.03/0.965} & \sd{36.48/0.972} & \sd{35.48/0.952} & \sd{34.54/0.969} & \sd{37.67/0.966} & \sd{33.71/0.959} & \sd{36.92/0.966} \\
\hline
\sa{DLADUN-9stg~\cite{liu2026degradation}} & \sd{37.72/0.967} & \sd{\underline{41.49}/0.982} & \sd{41.87/0.977} & \sd{46.14/0.990} & \sd{36.57/0.976} & \sd{\underline{38.10}/0.979} & \sd{37.23/0.962} & \sd{\textbf{36.43}/0.978} & \sd{40.85/0.979} & \sd{\underline{35.18}/\underline{0.969}} & \sd{39.16/0.976} \\
\Xhline{0.8pt}
\sa{\textbf{Ours-3stg}} & \sd{37.36/0.964} & \sd{39.94/0.976} & \sd{41.95/0.978} & \sd{46.67/0.991} & \sd{36.23/0.974} & \sd{36.94/0.975} & \sd{37.45/0.965} & \sd{35.42/0.974} & \sd{40.88/0.979} & \sd{34.49/0.963} & \sd{38.73/0.974} \\
\hline
\sa{\textbf{Ours-9stg}} & \sd{\textbf{38.42}/\underline{0.972}} & \sd{\textbf{41.76}/\textbf{0.984}} & \sd{\textbf{44.00}/\textbf{0.983}} & \sd{\underline{47.89}/\textbf{0.993}} & \sd{\textbf{38.47}/\underline{0.983}} & \sd{38.06/\underline{0.981}} & \sd{\textbf{39.37}/\underline{0.976}} & \sd{36.01/\textbf{0.979}} & \sd{\textbf{43.46}/\textbf{0.986}} & \sd{34.98/0.967} & \sd{\textbf{40.24}/\textbf{0.980}} \\
\Xhline{1.0pt}
\end{tabular}
\end{table*}

\begin{figure}[t]
  \centering
  \includegraphics[width=1.0\linewidth]{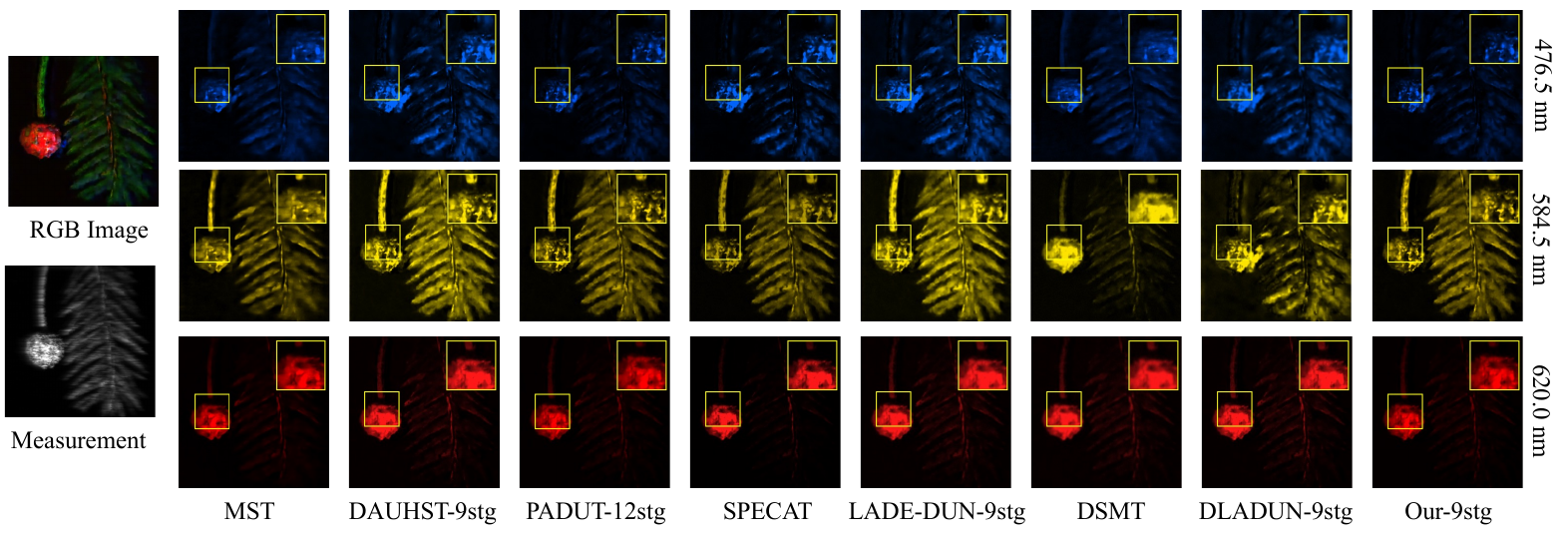}
  \caption{Qualitative comparisons of real HSI reconstruction on Scene 1 with 3 spectral channels.}
  \label{fig4}
\end{figure}

We further evaluate our model in real HSI reconstruction to validate its generalization capability beyond simulated settings. Following the same settings as~\cite{meng2020eccv,hu2022cvpr,huang2021cvpr} for a fair comparison, we re-train our model with the real mask on the CAVE and KAIST datasets. To simulate real imaging conditions, the training samples are injected with 11-bit shot noise. Fig.~\ref{fig4} shows the visual comparisons between our model and state-of-the-art methods across three representative spectral channels. Our model recovers finer structural details and cleaner textures with fewer artifacts, while other methods tend to produce blurry reconstructions or introduce noise-induced artifacts. These qualitative results further confirm that the manifold prior provides reliable structural guidance even under real measurement noise, demonstrating the robustness of the proposed framework.

\noindent\textbf{Computational Efficiency.}
Tab.~\ref{tab1} compares the theoretical complexity of methods in parameters and FLOPs. 
Our 9-stage model achieves the best average reconstruction performance with a moderate model size of 3.71M parameters and 70.62G FLOPs. 
Ours-3stg and Ours-9stg have identical parameter counts because the intermediate unfolding stages reuse a shared set of weights; increasing the number of stages therefore improves reconstruction quality without introducing additional parameter overhead. 
Although our 9-stage model has slightly more parameters than LADE-DUN-9stg, it requires fewer FLOPs (70.62G vs.\ 86.23G), indicating the improved computational efficiency of our Mamba-based unfolding design over the Transformer-based diffusion-prior counterpart. 
Moreover, under the same hardware setting, our method requires 132.76~ms runtime and 0.63~GB peak allocated memory, improving over LADE-DUN-9stg (140.15~ms and 0.65~GB) while achieving a 0.80\,dB gain in average PSNR. 
These results demonstrate a favorable accuracy--efficiency trade-off of our method.

\subsection{Ablation Study}

\textbf{Component Analysis.} To validate the contribution of each proposed component, we conduct an ablation study on the simulation dataset using a 3-stage model. As shown in Tab.~\ref{tab:ablation}, we progressively add each component to evaluate its impact on reconstruction performance. First, employing the Mamba denoiser as the backbone yields a baseline PSNR of 37.02\,dB, demonstrating its effectiveness in capturing long-range spatial-spectral dependencies with linear complexity. Building upon this, incorporating the Cross-Prior Attention module provides a further gain of 0.6\,dB over the Mamba baseline, which validates that cross-attention between image features and the manifold prior effectively injects global structural cues into local feature representations. Furthermore, the integration of the Prior-Mamba module yields a significant improvement of 1.05\,dB over the previous configuration, confirming the advantages of structure-aware scanning via manifold prior modulation. Finally, adding the Prior-Spectral Weight mechanism further enhances the results by 0.48\,dB, reaching a peak performance of 38.73\,dB and confirming that adaptive per-band modulation better handles spectral heterogeneity across wavelengths.

\begin{table}[htbp]
\centering
\caption{Ablation study on the CASSI simulation dataset.}
\label{tab:ablation}
\footnotesize
\begin{tabular}{ccccccc}
\Xhline{1.5pt}
Mamba  & Cross-Prior Attn & Prior-Mamba & Prior-Spectral Weight & PSNR (dB) & SSIM & Params (M) \\
\Xhline{1pt}
\checkmark  &   &   &    & 37.02 & 0.963   & 1.20 \\
\checkmark  & \checkmark  &      &     
& 37.62  & 0.968  & 3.17 \\
\checkmark  & \checkmark  & \checkmark  &   
&38.25     & 0.969    & 3.60 \\
\checkmark  & \checkmark  & \checkmark  & \checkmark  &
\textbf{38.73}  & \textbf{0.974} & 3.71 \\
\Xhline{1.5pt}
\end{tabular}
\end{table}

\begin{figure}[htbp]
    \centering
    \includegraphics[width=1\linewidth]{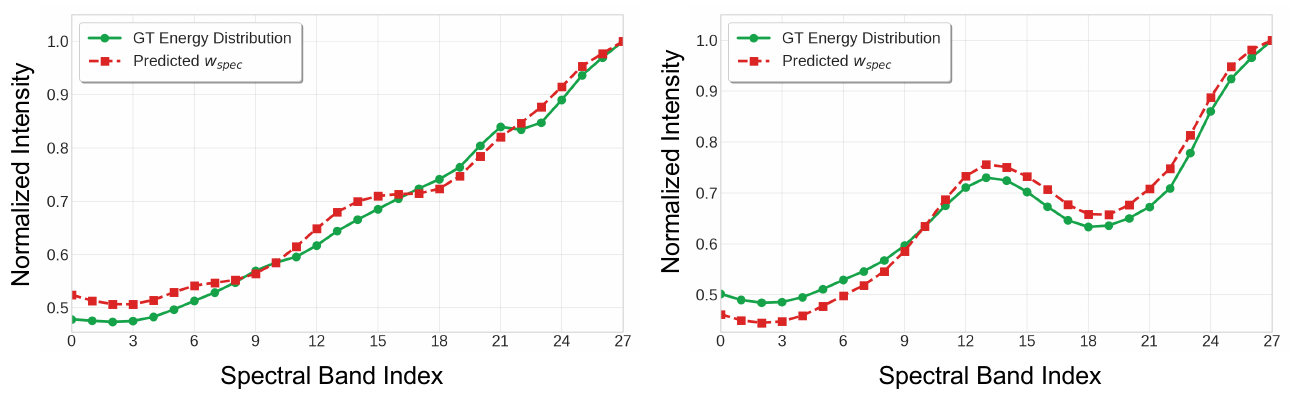}
    \caption{Comparison between the predicted prior spectral weights (red dashed) and the ground-truth spectral profile (green solid) on two datasets.}
    \label{fig5}
\end{figure}

\textbf{Effectiveness of Prior-Guided Spectral Weighting.} 
To verify that the learned prior effectively captures the spectral energy distribution of the input scene, we visualize the predicted spectral weights $\mathbf{w}_{\text{spec}}$ and compare them against the ground-truth spectral profile across two benchmark datasets. As shown in Fig.~\ref{fig5}, we randomly select one image from the CASSI dataset (left), and the optical filter dataset (right), and plot the ground-truth spectral profile (green solid line) and the predicted spectral weights (red dashed line) across all 28 spectral bands. The high agreement between the two curves demonstrates that the Prior-Guided Spectral Weight module successfully captures scene-specific spectral energy distributions, enabling the network to adaptively reweight the physical back-projection step across spectral bands and guide the unfolding reconstruction in a spectrally-aware manner.

\section{Conclusion}


In this paper, we propose a manifold-prior-guided deep unfolding framework that addresses the dual bottlenecks of scene-agnostic projection and artifact-contaminated denoising in HSI reconstruction. Our approach centers on a two-stage manifold learning strategy that distills a degradation-free structural prior, which is then synergistically injected into a prior-guided spectral weighting mechanism and a manifold-guided Mamba denoiser. By providing unified structural guidance with linear complexity, our method achieves state-of-the-art performance on both CASSI and optical filter-based systems while maintaining the interpretability of the physics-driven unfolding structure.

\section*{Acknowledgments}

This work was supported by the National Natural Science Foundation of China (62671049, 62506108 and 62331006), and Beijing Natural Science Foundation (L253009).

\newpage

\medskip
{
\small
}

\bibliographystyle{unsrt}
\bibliography{ref.bib}






\newpage
\section*{NeurIPS Paper Checklist}

\begin{enumerate}

\item {\bf Claims}
    \item[] Question: Do the main claims made in the abstract and introduction accurately reflect the paper's contributions and scope?
    \item[] Answer: \answerYes{}
    \item[] Justification: The abstract and introduction clearly state the two core contributions: (1) a two-stage manifold prior learning strategy injected into both the projection and denoising modules, and (2) a manifold-guided Mamba denoiser. These claims are fully supported by the experimental results in Section~3.

\item {\bf Limitations}
    \item[] Question: Does the paper discuss the limitations of the work performed by the authors?
    \item[] Answer: \answerYes{}
    \item[] Justification:Potential limitations of our method, including the increased training complexity of the two-stage strategy, and inference overhead from the diffusion model.

\item {\bf Theory assumptions and proofs}
    \item[] Question: For each theoretical result, does the paper provide the full set of assumptions and a complete (and correct) proof?
    \item[] Answer: \answerNA{}
    \item[] Justification: This paper does not include formal theoretical results or proofs. The proposed framework is derived from standard optimization theory (GAP) and variational inference, with all mathematical derivations presented inline in Section~3.

    \item {\bf Experimental result reproducibility}
    \item[] Question: Does the paper fully disclose all the information needed to reproduce the main experimental results of the paper to the extent that it affects the main claims and/or conclusions of the paper (regardless of whether the code and data are provided or not)?
    \item[] Answer: \answerYes{}
    \item[] Justification: All training details including optimizer, learning rate schedule,  number of epochs,  and two-phase training strategy are described in Section~4.1. The datasets used (CAVE and KAIST) are publicly available.

\item {\bf Open access to data and code}
    \item[] Question: Does the paper provide open access to the data and code, with sufficient instructions to faithfully reproduce the main experimental results, as described in supplemental material?
    \item[] Answer: \answerYes{}
    \item[] Justification: Link to GitHub will be available upon release of the paper.

\item {\bf Experimental setting/details}
    \item[] Question: Does the paper specify all the training and test details (e.g., data splits, hyperparameters, how they were chosen, type of optimizer) necessary to understand the results?
    \item[] Answer:\answerYes{}
    \item[] Justification: Section~3.1 provides complete experimental settings including datasets, evaluation metrics, optimizer hyperparameters, learning rate schedule, GPU hardware, 
and the two-phase training protocol.

\item {\bf Experiment statistical significance}
    \item[] Question: Does the paper report error bars suitably and correctly defined or other appropriate information about the statistical significance of the experiments?
    \item[] Answer: \answerYes{}
    \item[] Justification: The proposed method is evaluated on fixed benchmark datasets (CAVE and KAIST) following standard practice in HSI reconstruction literature. Quantitative results are reported in terms of PSNR and SSIM across 10 simulation scenes, providing sufficient statistical coverage to support the main claims. Consistent improvements are observed across all scenes and both imaging systems, confirming the statistical reliability of the reported results.

\item {\bf Experiments compute resources}
    \item[] Question: For each experiment, does the paper provide sufficient information on the computer resources (type of compute workers, memory, time of execution) needed to reproduce the experiments?
    \item[] Answer:\answerYes{}
    \item[] Justification: Section~4.1 specifies that all models are trained on a single NVIDIA RTX 3090 GPU. Training runs for 100 epochs with the described two-phase strategy.
    
\item {\bf Code of ethics}
    \item[] Question: Does the research conducted in the paper conform, in every respect, with the NeurIPS Code of Ethics \url{https://neurips.cc/public/EthicsGuidelines}?
    \item[] Answer:\answerYes{}
    \item[] Justification: This work focuses on hyperspectral image reconstruction for scientific and medical imaging applications. It does not involve human subjects, sensitive personal data, or any application with foreseeable negative societal impact.

\item {\bf Broader impacts}
    \item[] Question: Does the paper discuss both potential positive societal impacts and negative societal impacts of the work performed?
    \item[] Answer:\answerNA{}
    \item[] Justification: This work addresses a fundamental computational imaging problem with applications in remote sensing, medical imaging, and environmental monitoring. No foreseeable negative societal impacts are anticipated from improvements in hyperspectral image reconstruction.
    
\item {\bf Safeguards}
    \item[] Question: Does the paper describe safeguards that have been put in place for responsible release of data or models that have a high risk for misuse (e.g., pre-trained language models, image generators, or scraped datasets)?
    \item[] Answer: \answerNA{}
    \item[] Justification: The proposed method is a scientific imaging reconstruction framework. It does not involve generative models for content creation, personal data, or other high-risk assets that require special safeguards.

\item {\bf Licenses for existing assets}
    \item[] Question: Are the creators or original owners of assets (e.g., code, data, models), used in the paper, properly credited and are the license and terms of use explicitly mentioned and properly respected?
    \item[] Answer: \answerYes{}
    \item[] Justification: All datasets (CAVE, KAIST) and baseline methods are properly cited in the references. The CAVE and KAIST datasets are publicly available for research use.

\item {\bf New assets}
    \item[] Question: Are new assets introduced in the paper well documented and is the documentation provided alongside the assets?
    \item[] Answer: \answerYes{}
    \item[] Justification: The code for the proposed model will be released on GitHub upon publication of the paper, together with documentation and instructions for reproducing the main results.

\item {\bf Crowdsourcing and research with human subjects}
    \item[] Question: For crowdsourcing experiments and research with human subjects, does the paper include the full text of instructions given to participants and screenshots, if applicable, as well as details about compensation (if any)? 
    \item[] Answer:  \answerNA{}
    \item[] Justification: This paper does not involve any crowdsourcing or research with human subjects.

\item {\bf Institutional review board (IRB) approvals or equivalent for research with human subjects}
    \item[] Question: Does the paper describe potential risks incurred by study participants, whether such risks were disclosed to the subjects, and whether Institutional Review Board (IRB) approvals (or an equivalent approval/review based on the requirements of your country or institution) were obtained?
    \item[] Answer: \answerNA{}
    \item[] Justification: This paper does not involve human subjects and therefore does not require IRB approval.

\item {\bf Declaration of LLM usage}
    \item[] Question: Does the paper describe the usage of LLMs if it is an important, original, or non-standard component of the core methods in this research? Note that if the LLM is used only for writing, editing, or formatting purposes and does \emph{not} impact the core methodology, scientific rigor, or originality of the research, declaration is not required.
    \item[] Answer: \answerNA{}
    \item[] Justification: LLMs are not used as any component of the core methodology. They were used only for writing assistance and formatting purposes, which does not require declaration per NeurIPS policy.

\end{enumerate}

\end{document}